\pdfoutput=1
\documentclass[11pt]{article}
\usepackage[]{EMNLP2022}
\usepackage{amsmath}
\usepackage{graphicx}
\usepackage{epstopdf} 
\usepackage{multirow}
\usepackage{times}
\usepackage{latexsym}
\usepackage{amssymb}
\usepackage{graphicx}
\usepackage[T1]{fontenc}
\usepackage{hyperref} 

\usepackage{subfigure}
\usepackage{parskip}
\usepackage{bm}
\usepackage[utf8]{inputenc}
\usepackage{hyperref}
\usepackage{makecell}
\usepackage{microtype}
\usepackage[lined,boxed,commentsnumbered]{algorithm2e}
\usepackage{inconsolata}
\usepackage{booktabs}

\title{How far is Language Model from 100\% Few-shot Named Entity Recognition in Medical Domain}
\author{Mingchen Li, \textbf{Rui Zhang} \\
        University of Minnesota, Twin Cities \\ \{li003378, zhan1386\}@umn.edu \\ 
        }
          
\begin{document}
\maketitle

\begin{abstract}

Recent advancements in language models (LMs) have led to the emergence of powerful models such as Small LMs\footnote{We define SLMs as pre-trained models with fewer parameters compared to models like GPT-3/3.5/4, such as T5, BERT, and others.} (e.g., T5) and  Large LMs (e.g., GPT-4). These models have demonstrated exceptional capabilities across a wide range of tasks, such as name entity recognition (NER) in the general domain.
Nevertheless, their efficacy in the medical section remains uncertain and the performance of medical NER always needs high accuracy because of the particularity of the field. This paper aims to provide a thorough investigation to compare  the performance of LMs  in medical few-shot NER and answer How far is LMs from 100\% Few-shot NER in Medical Domain, and moreover to explore an effective entity recognizer to help improve the NER performance.  
%%%%%%%
Based on our extensive experiments conducted on 16 NER models spanning from 2018 to 2023, our findings clearly indicate that LLMs outperform SLMs in few-shot medical NER tasks, given the presence of suitable examples and appropriate logical frameworks. Despite the overall superiority of LLMs in few-shot medical NER tasks, it is important to note that they still encounter some challenges, such as misidentification, wrong template prediction, etc.
%%%%%%%%%%%%%
Building on previous findings, we introduce a simple and effective method called \textsc{RT} (Retrieving and Thinking), which serves as retrievers, finding relevant examples, and as thinkers, employing a step-by-step reasoning process.
Experimental results show that our proposed \textsc{RT} framework significantly outperforms the strong open baselines on the two open medical benchmark datasets\footnote{The source code, data are available at \url{https://github.com/ToneLi/RT-Retrieving-and-Thinking}.}.

% few works explore the importance of example selection in the NER task. most work  explores text classification.

% 1) what is a good demonstration?

% 2)how to select the ner examples?

% 3) mask method is not useful

% 4) more examples is useful or not?

% 5)  the action must be same as the instruction says
% 5) 99.06 

\end{abstract}

\section{Introduction}

Researchers are increasingly interested in applying information extraction to mine a vast quantity of unstructured information from electronic medical records.
These techniques can offer valuable perception  and generate substantial benefits  for 
clinical research, such as drug discovery~\cite{zhang2021drug}, knowledge graphs building~\cite{li2020multi,wu2023medical}, question answering~\cite{li2022semantic,pugachev2023consumer} and link prediction~\cite{li2022hierarchical,zheng2023sprda}. Within the scope of medical text mining, one of the  most essential tasks in medical text mining is medical named entity recognition (NER). However,
 existing supervised medical NER models necessitate a substantial amount of human-annotated (i.e. medical student, doctor) data. To tackle this issue, few-shot techniques have been introduced to perform NER in resource-constrained settings by leveraging auxiliary information or improving the discrimination between different labels.

 To address the challenges posed by the scarcity of available medical data, recent studies propose to harness the power of  SLMs or LLMs.  When designing models based on SLMs, the current methods primarily utilize  nearest neighbor inference~\cite{yang2020simple,das2022container,ji2022few}, prompt tuning~\cite{huang2022copner,liu2022qaner}, contrastive learning~\cite{zhang2022optimizing,das2022container,li2023w} or generation methods~\cite{li2023understand,wang2023gpt}.  On the other hand, when designing the models based on LLMs, the main approach is to employ the in-context learning (ICL)~\cite{min2022rethinking} method to stimulate the recognition ability of LLMs. 
%%%%%%%%%%%%%
 Despite the impressive performance showcased by current works, there remains a lack of comprehensive investigation comparing the performance of LMs in medical few-shot NER. Otherwise, we found researcher tends to design the models to solve the NER problems in the general domain, become of the issues of secret data, even though these models can get high performance. More important, The NER task in the medical domain demand near-perfect (100\%) accuracy due to their critical implications for human life. Consequently, determining the suitability of these models for the medical domain presents a challenging problem.

In this paper, our objective is to conduct a comprehensive evaluation of the advantage and disadvantages of SLMs and LLMs on the medical  few-shot NER tasks, and then answer the following questions: \textbf{1)}  Which of the two, SLMs or LLMs performs better in the few-shot medical NER task? \textbf{2)} Is it possible to transfer NER models trained for the general domain to the medical domain? \textbf{3)} Do the quantity and quality of annotation have any impact on the performance of SLMs and LLMs? \textbf{4)} How far is LM from 100\% few-shot  medical NER?

To answer these questions, we conduct an extensive empirical study involving 16 different few-shot NER models spanning from 2018 to 2023.  Our study specifically focuses on evaluating these models using two well-established open medical NER datasets, ensuring a standardized and comprehensive analysis.   
%%%%%%%%%%%%%%%%% 
The results show that 1)  LLMs demonstrate superior performance over SLMs when the LLMs are provided with high-quality instructions. 
%%%%%%%%%%%%%%%%%%%%%%%%%%%%%
2)  Our findings indicate that in order to successfully transfer a NER model from the general domain to the medical domain, it is essential to pre-train the NER models specifically in the medical domain.  The NER in the medical domain  presents the same challenges, including flat/net entity recognition, abbreviation handling, and recognizing long-word entities, etc.  The pre-training enables the model to acquire the necessary prior knowledge and context relevant to medical entities.  
%%%%%%%%%%%%%%%%%%%%%%%%%%%%%%%%
3) The quantity of samples has a greater impact on SLMs compared to LLMs. Additionally, when we examine the scenario where the training data is only partially annotated, we observe that LLMs exhibit more consistent and stable performance compared to SLMs.   Regarding the aspect of quality, our findings contradict certain studies that suggest improved performance for LLMs with retrieved relevant samples instead of using random samples. Our results indicate that the effectiveness of this approach varies depending on the specific datasets employed in the evaluation. Otherwise,  we found the effectiveness of LLMs is greatly influenced by the careful selection of appropriate examples and the application of sound entity recognition logic.  In our latest findings, we found the choice between using a combination of positive and negative examples or exclusively positive examples has a definite influence on the overall performance of LLMs with ICL.
%%%%%%%%%%%%%%%%%%%%%
4) We have identified several  factors that contribute to the inability of the current state-of-the-art model to achieve 100\% entity recognition. One reason is that NER models face challenges in extracting long, special entities and Out-of-vocabulary  entities. Furthermore, the superior performance of LLMs is highly reliant on pre-existing knowledge and contextual understanding, the failure to timely update the knowledge  can have a significant impact on the overall performance and effectiveness of the entity recognition system. Otherwise, there are many entities annotated in the training or testing dataset has more than one entity type. For the full analyses, please check Section.\ref{Error Analysis}.

Based on some significant findings, we have proposed a novel few-shot entity extractor called  RT (\textbf{R}etrieving and \textbf{T}hinking). The main focus of RT is to consider both the selection of appropriate examples  and effective entity recognition logic. The basic idea of RT involves utilizing a basic LLM to identify the entity classes present in a given sentence. Subsequently, we employ K-nearest neighbors (KNN) to retrieve relevant examples for the sentences based on the recognized entity classes. Finally, following the approach suggested by ~\cite{wei2022chain}, we design a logical entity recognition process that enables the model to identify entities in a step-by-step manner.
%%%%%%%%%%%%
We perform extensive experiments on two standard medical datasets and 16 NER models for few-shot  medical NER demonstrating the superiority of our method over prior state-of-the-art methods. Our contributions are the following:

\begin{itemize}
    \item We conduct an extensive empirical study comparing SLMs and LLMs on medical few-shot NER tasks across 16 NER models during 2018-2023.
\item  We  propose \textsc{RT}, a new framework that uses Retrieving and Thinking strategies to improve the performance of LLMs with ICLs on medical few-shot NER tasks.
\item  We conduct a thorough analysis of our method, including an ablation study, demonstrating the ineffectiveness of  \textsc{RT}.
\end{itemize}

\section{Related Work}

\subsection{Small Language Models (SLMs) in Medical Few-Shot NER}
Few-shot named entity recognition is a task that aims to predict the label (type) of an entity from insufficient labeled data. Most previous work in the medical domain prefer to directly use medical language models such as ClinicalBERT~\cite{huang2019clinicalbert}, BioBERT~\cite{lee2020biobert}, and GatorTron~\cite{yang2022large}  to solve the few-shot NER problem.  A few studies
\cite{fritzler2019few, ji2022few} propose to utilize the prototype network \cite{snell2017prototypical} to catch the few-shot medical NER tasks. 
%%%%%%%%%%
Inspired by the  nearest neighbor inference~\cite{wiseman2019label}, \cite{yang2020simple} proposes NNshot and Structshot, which use the nearest neighbor to search for the  nearest label of each testing entity.
% and then use the Viterbi Decoder to capture label dependencies. 
In CONTaiNER~\cite{das2021container}, the authors use contrastive learning to increase the  discrimination for each label and adopt Gaussian embeddings for each token to solve the Anisotropic  property in the few-shot NER task. 
The prompt-based method is also explored in this task, such as \cite{huang2022copner} uses the prompt to guide contrastive learning. 
\cite{zhang2022optimizing} proposes a span-based contrastive learning method to handle the nested NER.  In W-Procer~\cite{li2023w}, the authors use a weighted prototypical contrastive learning method to solve the class collision issue in few-shot medical NER.  MetaNER~\cite{chen2023learning}  proposes a novel approach to enhance the performance of NER by pre-training SLMs using instructions and demonstrations. As the popularity of LLMs continues to rise, it becomes crucial to assess the effectiveness of SLMs, so in our work, we provide a thorough investigation to compare the performance of SLMs and LLMs in the medical few-shot NER.

\subsection{Large Language Models (LLMs) in Medical Few-Shot NER}
There are few works to explore the effectiveness of LLMs in Few-Shot NER, so, in this section, we present relevant works exploring the use of Large Language Models (LLMs) in the general domain. Furthermore, we evaluate the performance of NER models using LLMs specifically in the medical domain.
%%%%%%%%%%%
% GPT-NER~\cite{wang2023gpt} proposes a method by retrieving the samples using the KNN method to improve the performance of in-context learning. The idea of the chain of thought is used in NER in PromptNER~\cite{ashok2023promptner}.  
\cite{ma2023large}  introduces a filter-then-rank  method by combing the advantage of SLMs and LLMs. The SLM is employed to filter the candidate labels for each token, while the instruction method is utilized to predict the final answer, which is performed by the LLM.
%%%%%%%%%%%%%%%%%%%
GPT-NER~\cite{wang2023gpt} presents a novel approach that enhances in-context learning performance by employing entity-level embedding and self-verification. PromptNER~\cite{ashok2023promptner} introduces the concept of a chain of thought into Named Entity Recognition. In our work, to improve the medical few-shot NER performance, we propose a method by retrieving the good samples and providing the right entity recognition logic.

% In another study by~\cite{ma2023large}, a filter-then-rank approach is proposed that combines the strengths of Small Language Models (SLMs) and Large Language Models (LLMs).
% The authors utilize SLMs to filter candidate labels for each entity and employ instruction-based methods to predict the final label, a step that is executed using LLMs.

\section{Small LMs v.s. Large LMs}
In this section, we aim to provide a thorough investigation to compare
the performance of SLMs and LLMs in medical few-shot NER and answer How far are LMs from 100\%
Few-shot NER in Medical Domain. To this end, we evaluate 16 NER models on two standard few-shot NER datasets.

\subsection{Problem Statement and Setups}
\label{con:task defination}
In this section, we formalize the task of  few-shot named entity recognition (NER) and present a standardized evaluation setup to ensure a fair comparison against previous SOTA models.

\subsubsection{Few-Shot NER}
In the traditional named entity recognition, given an input sentence of $i$ tokens $\boldsymbol{x}=\{x_1,x_2,...,x_i\}$, the NER model intends to assign each token $x_i$ to its corresponding label $y_i$.  In the few-shot NER task, we assume that there are two same label sets $\{L^n\}$ of $n$ length in the support set (training set) and query set (testing set). The $K$-shot NER task is formally defined as follows: given the input sentence  $\boldsymbol{x}$ and a set of $K$-shot entities for each label class in  $\{L^n\}$, it aims to find the best sequence $\boldsymbol{y}=\{y_1,y_2,...,y_i\}$ for $\boldsymbol{x}$. $K$-shot entities set contains $K$ entity examples for each entity class in $\{L^n\}$.

\subsubsection{Setups}
For the evaluation scheme on few-shot NER, the current work~\cite{fritzler2019few,hou2020few} adapted the episode evaluation method which just utilizes episode data from the test set. As shown in \cite{das2021container}, sampling the test episodes from the actual test set perturbs the real distribution of testing data that may not reflect the actual performance. So, same as the  existing studies~\cite{das2021container,yang2020simple,huang2022copner} about Few-shot NER, 
we employ the original test set of NCBI for our prediction on LLMs.
%%%%%%%%%%%%%%%%%%%
Due to the expensive nature of the GPT API, we have chosen to sample 100 instances from the BC5CDR dataset as the test set for the LLMs. We randomly selected one LLM to evaluate the LLM has a similar performance under the whole test set and sampled 100 instances on  BC5CDR.
However, for the SLMs, we continue to utilize the original test set of BC5CDR and NCBI.
%%%%%%%%%%%%%%%%%%%%%%%%%
we utilize greedy sampling~\cite{yang2020simple} to sample  the support set. 
To clarify the tagging scheme setup, we utilize the \textsc{"IO"} tagging scheme. Under this scheme, the \textsc{"I"}  indicates that all tokens are inside an entity, while \textsc{"O"} refers to all other tokens.

\subsection{Dataset}
\begin{table}[ht]
	\centering
	
	\renewcommand\arraystretch{1.3}
	\scalebox{0.7}{
	\begin{tabular} {cccc}
		\hline 
		
		\hline	
		Dataset& \# Domain & \# Class &   \# Sentence/ \# Entity\\ 
		\hline		
		% I2B2'14&Medical &  23& 140,817/\,29,233\\
		BC5CDR&Medical& 2 &13,938/\,28,545\\
        NCBI&Medical& 4&    7,287/\,7,025\\
		\hline
	\end{tabular}}
		\caption{ Datasets Statistics. \# Class refers to the number of entity classes (types) that have been labeled in a dataset.}
	\label{con:FENR data}
\end{table}

SLMs and LLMs are evaluated with two medical datasets, including BC5CDR~\cite{li2016biocreative} and NCBI~\cite{dougan2014ncbi}. Due to concerns arising from the I2B2~\cite{stubbs2015annotating} user agreement, we refrained from conducting a comparison of LLMs within the I2B2 dataset.
Among these datasets,  NCBI and BC5CDR consist of 798 and 1500 public medical abstracts separately, all of which are annotated with MeSH identifiers. 
Table~\ref{con:FENR data} shows the detailed statistical information of these three datasets.

\subsection{SLMs, LLMs and Evaluation Metrics}
\subsubsection{Small Language Models}
We  provide a thorough investigation  with several strong baselines based on the state-of-the-art pre-trained few-shot NER models, including both Domain Transfer and Domain non-Transfer methods. Specifically, the Domain Transfer models have been trained on the OntoNotes 5.0 dataset, which comprises medical NER knowledge. The evaluation of these models is conducted using the support sets and test sets from BC5CDR, and NCBI.  On the contrary, Domain non-Transfer models differ in that they do not undergo additional pre-training using other medical NER datasets.

we consider the following Domain Transfer models:  (1) \textbf{NNshot}\cite{yang2020simple} is a method that uses nearest neighbor classification.
(2) \textbf{Structshot}\cite{yang2020simple} is an improved version of NNshot that combines nearest neighbor classification, abstract transition matrix, and Viterbi algorithm. (3) \textbf{ContaiNER}~\cite{das2022container}  adopts contrastive learning to estimate the distributional distance between entities' vectors, which are represented using Gaussian embeddings. (4) \textbf{COPNER}~\cite{huang2022copner} leverages contrastive learning with prompt tuning to  identify entities.  (5) \textbf{EP-NET}~\cite{ji2022few} is a NER method based on the dispersedly distributed prototypes network.  (6)  \textbf{MetaNER}~\cite{chen2023learning} proposes a novel approach to enhance the performance
of NER by pre-training SLMs using instructions
and demonstration.

we also consider the following Domain non-Transfer models: (1) \textbf{ProtoBERT}~\cite{huang2022copner} is a few-shot NER method, which utilizes the prototypical network~\cite{snell2017prototypical} and BERT model to infer the entity label. (2) \textbf{BINDER}~\cite{zhang2022optimizing} employs span-based contrastive learning to effectively push the entities of different types by optimizing both the entity type encoder and sentence encoder.
(3) \textbf{LM-tagger} (BERT~\cite{devlin2018bert}, ClinicalBERT~\cite{huang2019clinicalbert}, BioBERT~\cite{lee2020biobert} and GatorTron~\cite{yang2022large}) are traditional SLM-based methods  that fine-tune the SLM on the support set with the label classifier.
%%%%%%%%%%%%%%%
Same as ~\cite{yang2020simple,das2022container,huang2022copner,ji2022few,zhang2022optimizing}, we evaluate all the models based on the generative evaluation metric, Micro F1.

\begin{table*}[ht]
	\centering
 \renewcommand\arraystretch{1.3}
\resizebox{0.7\textwidth}{!}{%
	\begin{tabular} {l|l | cc |cc}
		\toprule 
		\multicolumn{1}{c}  {}&\multicolumn{1}{c}  {}& \multicolumn{2}{c}  {1-shot}& \multicolumn{2}{c}  {5-shot}\\
		&Approach &BC5CDR &NCBI & BC5CDR&NCBI   \\ 
          \midrule
        
          % SimBERT-NN~\cite{yang2020simple}  &  7.7$\pm$0.8 &&   &  9.1$\pm$0.7 &  &\\
         % BERT-NN & \textbf{7.6$\pm$0.45}  & &   &\textbf{8.5$\pm$0.55} && \\
         \multirow{12}*{SLMs}&BERT~\cite{devlin2018bert}    &10.58   &  12.33  & 35.60  & 26.532\\
         ~&ClinicalBERT~\cite{huang2019clinicalbert} &  19.96  & 8.27  & 39.25 & 24.47\\ % from allennlp 
        ~&NNShot~\cite{yang2020simple}   &   32.96 & 11.82  &39.30 & 16.22 \\
        ~&BioBERT~\cite{lee2020biobert}   & 36.78   &  27.19  & 47.04&  35.88 \\
         ~&StructShot~\cite{yang2020simple}    &16.09& 4.63 &30.97& 13.89 \\
          ~&ContaiNER~\cite{das2022container}    & 37.25 &16.51  &   41.21  & 26.83\\
          ~&COPNER~\cite{huang2022copner}  &36.36 &  15.54  &42.78 & 24.23\\
        ~&ProtoBERT~\cite{huang2022copner}   &23.61 &  17.24&  40.58&34.18\\
        ~&GatorTron~\cite{yang2022large}     & 26.97  & 35.00  &55.44 &37.64 \\
            ~&BINDER~\cite{zhang2023optimizing}  &1.86&  2.95  & 51.79 & 31.95\\
         ~&\textsc{W-Procer}~\cite{li2023w}    & 40.26   &38.86&  56.02 & 40.90 \\
     ~&MetaNER~\cite{chen2023learning}&     -- &   40.01  &--  & 44.92   \\
        \midrule
       \multirow{4}*{LLMs}&Vanilla ICL~\cite{li2023w}  &  43.90 &43.48     &44.72   &  43.83 \\
         
          % ~&GPT-3.5-turbo  & 30.37$\pm$1.26 &  22.60$\pm$1.18   &   39.92$\pm$1.19  & 33.64$\pm$0.55  \\
         ~&PromptNER~\cite{ashok2023promptner}&   92.84 &   81.29 &    93.33  & 84.06  \\
          % ~&PromptNER  (just right)&  91.85   &     &      &   \\
           ~&GPT-NER*~\cite{wang2023gpt}&  91.28  &  90.44   &  91.56  & 90.44  \\
       ~&GPT-NER*(self-ve)~\cite{wang2023gpt}& 91.28   &  90.79  & 91.63  &   90.67 \\
        %  ~&PromptNER (mask1)&    92.39  &   82.71   &       &       \\
        %  \midrule
        %  ~&PromptNER (tree-of-thought)&    91.50  & -- &       &       \\
        %  \midrule
        %   ~&PromptNER (assume know all entity)&    \textbf{99.06}  & \textbf{ 95.92 }   &   \textbf{99.06}     &   \textbf{95.92}     \\
        %   % \midrule
       
        %  ~&GPT-NER (chain of thought just right excamples)&    92.38    &       &     &    \\
        % ~&GPT-NER (chain of thought wrong/right excamples)&   93.72    &       &     &    \\
        % % ~&GPT-NER (chain of thought wrong/right excamples-add-prompt-example)&   93.47   &       &     &    \\
        %     % ~&GPT-NER (chain of thought wrong/right excamples-no-add-prompt-example)&   92.57 &       &     &    \\
        % ~&Our-method&   93.50 &   91.56  &  93.26 (6.31)  &  91.76 (9.46)  \\
        % % ~&Our-method-mask&    &      &     &    \\
        % ~&Our-method-just-right-excample&  --  &    --  &     &    \\
        % ~&Our-method-hun-excamples&  --  & --     &     &    \\
       
         % ~&Filter-then-Rerank~\cite{ma2023large}&     -- &  -- &  &    \\
        \bottomrule            
	\end{tabular}
 }
  \caption{ The performance comparison of SLMs and LLMs in BC5CDR, and NCBI. self-ve refers to self-verification.
  }
  \vspace{+2mm}
\label{con:Model_performance}
\end{table*}

\subsubsection{Large Language Models}
We adapt the current strongest GPT-4\footnote{https://platform.openai.com/docs/models/gpt-4} rather than GPT-3.5\footnote{https://platform.openai.com/docs/models/gpt-3-5}  in our experiments out of two primary reasons: (1) According to recent findings by ~\cite{li2023w}, GPT-4 demonstrates superior performance in few-shot medical named entity recognition (NER) tasks compared to GPT-3.5. (2) GPT-4 is equipped with a more extensive set of pre-existing knowledge and a larger number of training parameters compared to its predecessor, GPT-3.5. By employing the respective source code, we have implemented a set of robust few-shot named entity recognition (NER) methods on the BC5CDR and NCBI datasets. All of these methods are specifically designed based on in-context learning principles, and GPT-4. \textbf{Note that}: The cost of running these models typically ranges from 6\$ to 10\$ per run by  using the GPT-4.

1) \textbf{Vanilla ICL}~\cite{li2023w} utilizes a set of common prompts that include instructions and demonstrations (examples). These examples are generated by randomly retrieving demonstrations from a database.
2) \textbf{PromptNER}~\cite{ashok2023promptner} follows a similar sample retrieval strategy as Vanilla ICL~\cite{li2023w}. However, PromptNER incorporates the concept of a chain of thought to enhance entity recognition.
3) \textbf{GPT-NER*}.  GPT-NER~\cite{wang2023gpt} uses a method by retrieving the samples using the KNN method to improve the performance of in-context learning.  In the original GPT-NER implementation, the authors did not provide distinct symbols for different labels. To address this, we introduced special symbols in the output of our modified version called GPT-NER*. We give an example in Figure~\ref{con:output_comparision}.
4)  \textbf{GPT-NER* (self-verification)}~\cite{wang2023gpt} follows GPT-NER* and adopts a new strategy to verify the accuracy of the recognition.

Same as ~\cite{yang2020simple,das2022container,huang2022copner,ji2022few,zhang2022optimizing}, we evaluate all the models based on the generative evaluation metric, Micro F1.

\section{Comparison Results}

We evaluate 16 approaches, 12 SLM-base NER models, and 4 LLM-based NER models on 2 open medical NER datasets. We first conduct pivot experiments and observe LLMs without  good instructions will reduce the performance of NER. As indicated in Table~\ref{con:Model_performance}, the Vanilla ICL exhibit lower performance compared to MetaNER and W-PROCER on the NCBI 5-shot and BC5CDR 5-shot datasets, respectively. Conversely, PromptNER and GPT-NER achieve exceptionally high performance on these datasets when compared to the results of SLMs.

\textbf{Can we transfer the NER models from the general domain to the medical domain.} The utilization of BERT models trained on general encyclopedic data poses challenges in accurately identifying medical entities within the medical domain. Despite various research efforts to redesign NER models like NNshot and StructShot, traditional SLM in the general domain still struggle to achieve ideal results in this context. However, pre-training models specifically on medical datasets, such as BioBERT, have shown promising performance gains compared to traditional BERT models. Consequently, researchers have focused on enhancing medical NER performance by pre-training models in the medical domain, as observed in projects like GatorTron and MetaNER. This suggests that equipping language models with domain-specific prior knowledge is crucial for improving NER performance.

\textbf{SLMs are more susceptible to the amount of training data.} From Table~\ref{con:Model_performance}, we can see that the quantity of samples has a greater impact on SLMs compared to LLMs. As an illustration, consider GPT-NER, which demonstrates comparable performance on both the NCBI 1-shot and 5-shot datasets. However, in the case of SLMs like MetaMER and ProtoBERT, the performance on the 5-shot dataset significantly surpasses that of the model performance on the 1-shot dataset. It shows that the SLMs are more susceptible to the amount of training data.

\textbf{Handling problems related to partially annotated datasets is comparatively easier for LLM.} 
\begin{figure}[htbp]
	\centering
	\includegraphics[width=0.8\columnwidth]{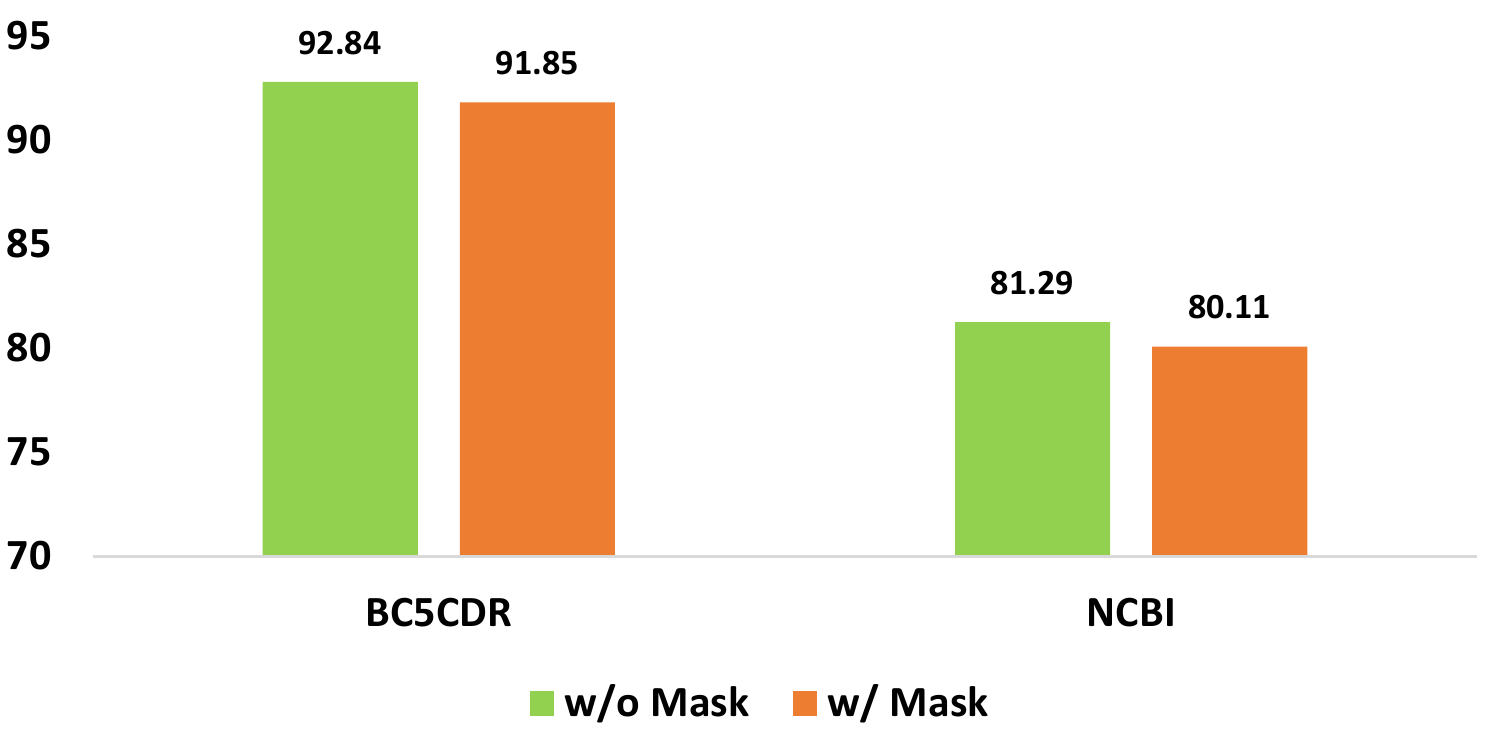} 
	\caption{LLM performance on without (w/o) mask operation and with (w/) mask operation}.
	\label{con:mask}
\end{figure}

In the study conducted by \cite{li2023w}, they experimented by masking certain labeled entities to simulate a partially annotated dataset. The authors observed a 3-4 point decrease in NER performance before and after the mask operation. So  in this paper, to test the ability of LLM, we mask the labeled entities on the 5-shot dataset, to make sure each label just has one entity, and then compare the performance of LLMs on the masked dataset and source 1-shot dataset. We use PromptNER in this experiment, and on the dataset BC5CDR and NCBI. The results are shown in Figure.~\ref{con:mask}. The experiment revealed that the partially annotated dataset had a minimal impact on the LLM.

\section{More Analysis on LLMs}
\begin{figure}[htbp]
	\centering
	\includegraphics[width=1.0\columnwidth]{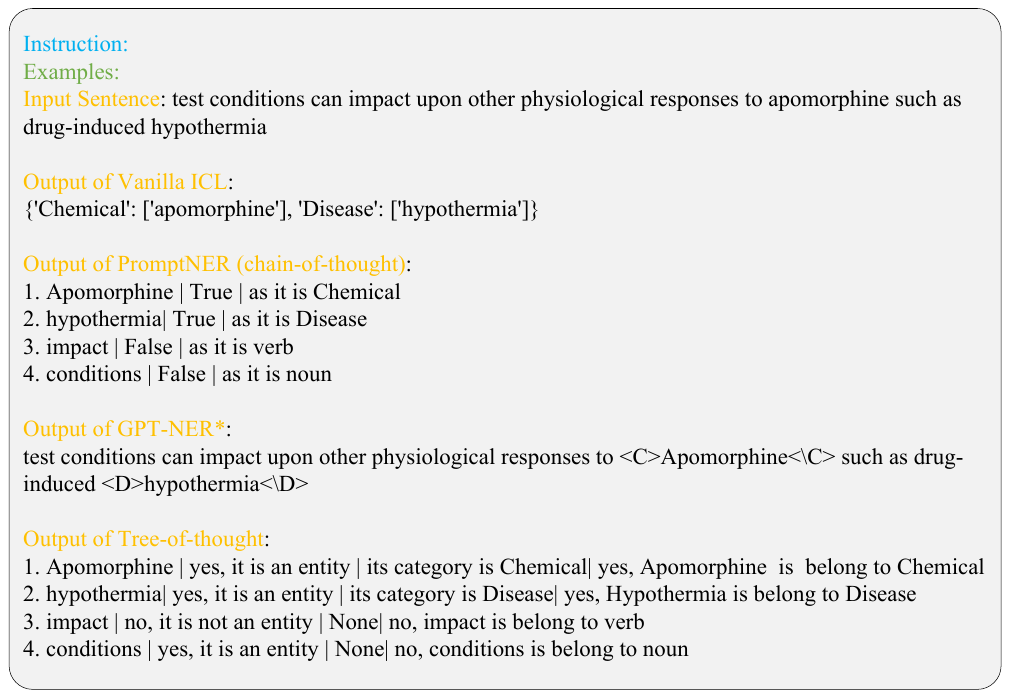} 
	\caption{The output format of different ICL methods}
	\label{con:output_comparision}
\end{figure}
\subsection{Positive and Negative Samples in the Chain-of-Thought}

\begin{figure}[htbp]
	\centering
	\includegraphics[width=0.8\columnwidth]{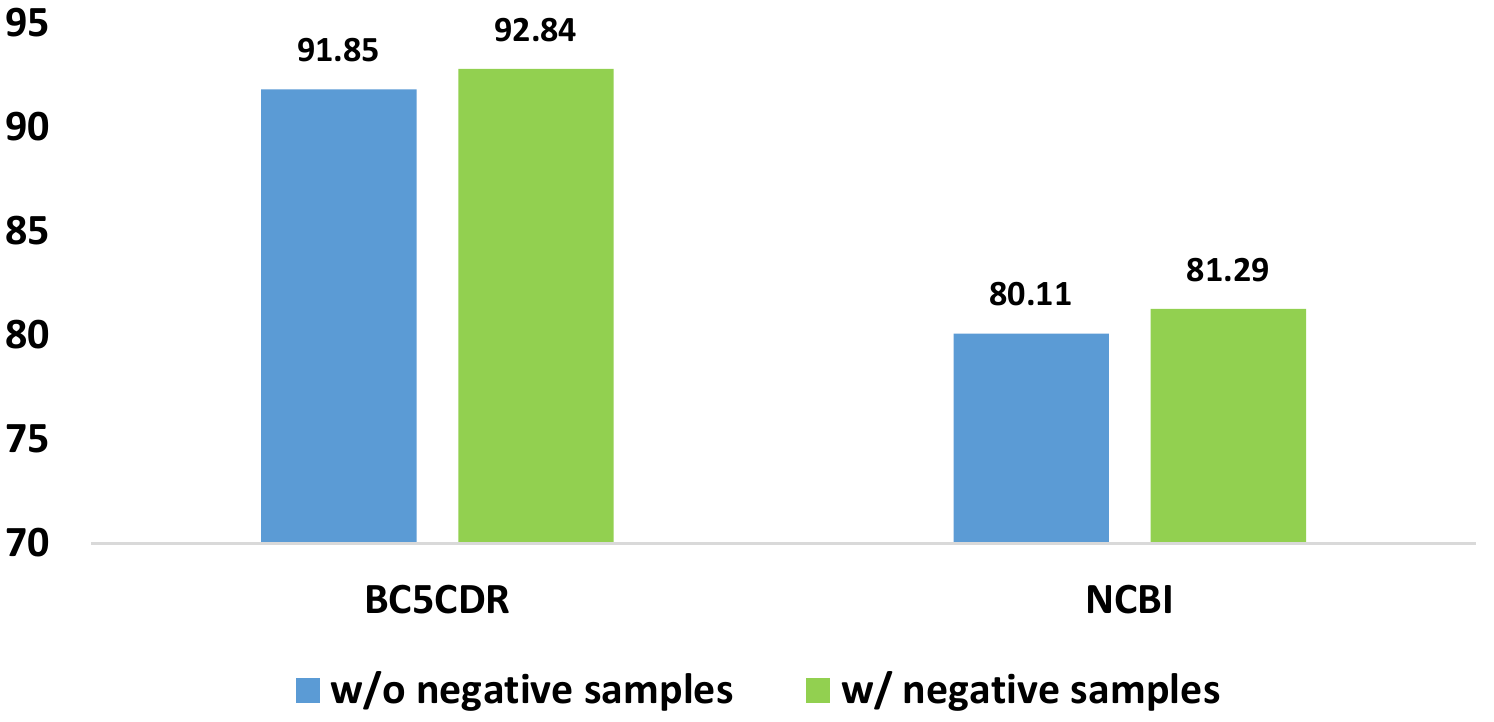} 
	\caption{LLM performance on without (w/o) negative samples and with (w/) negative samples}
	\label{con:negative_samples}
\end{figure}
Certain LLMs, such as PromptNER, which emphasizes the utilization of the chain of thought in the NER task, incorporate a design that includes both positive and negative samples within the instructions.  As shown in Figure~\ref{con:output_comparision}, the positive example is "Apomorphine | True | as it is Chemical" and the negative example is "impact | False | as it is a verb".
It is important to evaluate the necessary of using the negative samples. In this work, we set a  simple experiment by comparing the model performance with the model that just uses the positive samples. As shown in Figure~\ref{con:negative_samples}, 
the results show that the performance of the model using the negative sample and positive samples is better than the model just using the positive samples.
%%%%%%%%%%
However, this evaluation is based on the assumption that the model solely employs a chain-of-thought strategy to enhance its generation output. In more intricate scenarios, further examination and discussion will be discussed in the future.
\subsection{The Influence of Different Output Examples}

\begin{figure}[htbp]
    \centering
    \subfigure[]{
        \includegraphics[width=1.42in]{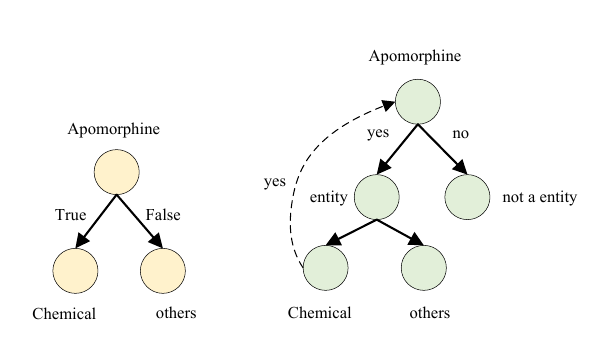}
    }
    \subfigure[]{
	\includegraphics[width=1.42in]{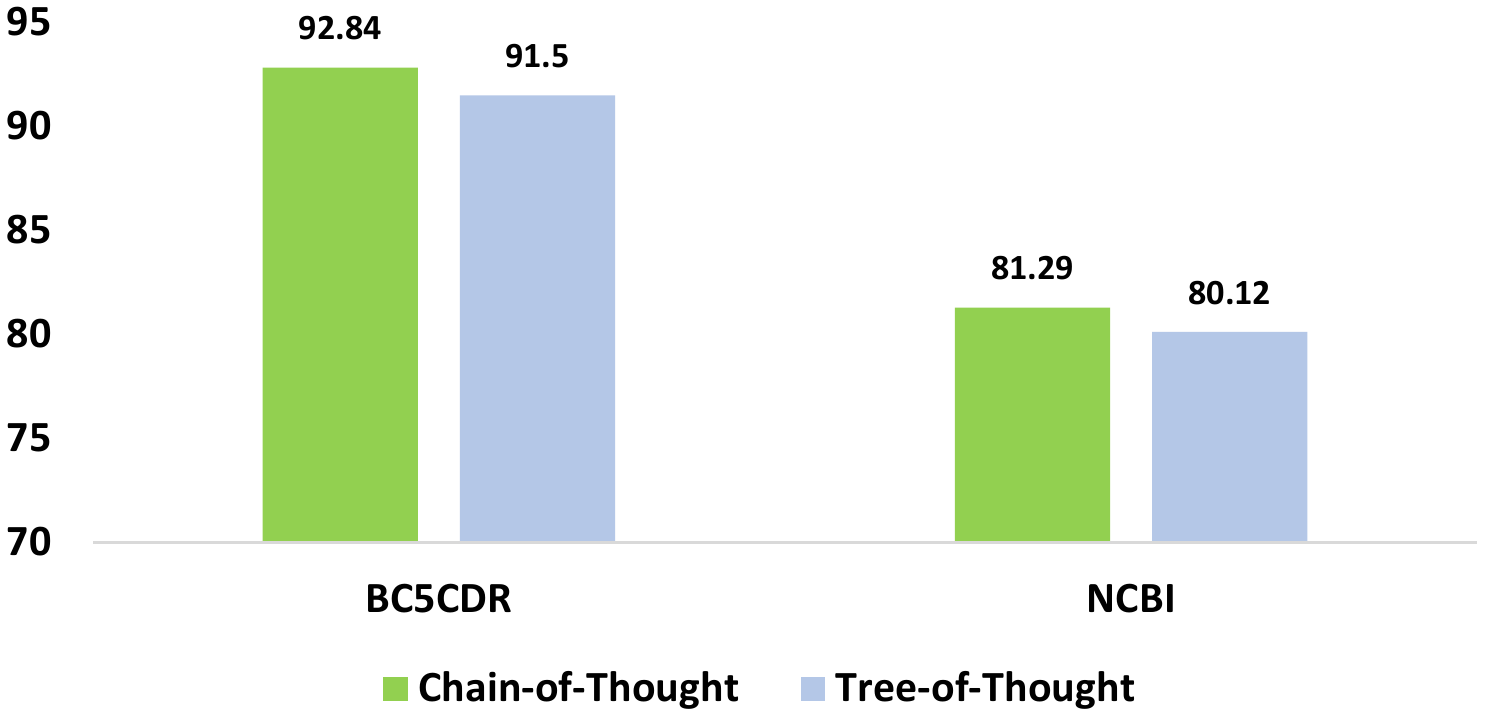}
    }
    \caption{(a) refers to the comparison of the chain of thought and tree of thought. (b) The LLM performance of the chain of thought and tree of thought.}
    \label{tree_chain}
\end{figure}

In Figure.~\ref{con:output_comparision}, we give a detailed example of output on different ICL strategies, Vanilla ICL, the chain of thought, and the tree of thought~\cite{yao2023tree}. For the idea of the tree of thought, we design the examples (samples) in the instruction as shown on the left of Figure~\ref{con:output_comparision}.
We have identified that the reason for the poorer performance on Vanilla ICL is the model's difficulty in predicting the dictionary symbol, such as "$]$", and"$\}$".
%%%%%%%%
By the comparison result in  Figure~\ref{tree_chain} and Table~\ref{con:Model_performance}, we found the chain of thought has a better performance than the tree of thought. The reason is that the tree of thought is more suitable   for the complex reasoning task, such as mathematical calculation. The impressive results achieved by PromptNER and GPT-NER also highlight the superiority of the chain of thought and KNN retrieval method in the context of medical few-shot NER tasks.

\subsection{LLMs are good for labeling entities, but not good for extracting entities}
\begin{table}[ht]
	\centering
 \renewcommand\arraystretch{1.3}
\resizebox{0.4\textwidth}{!}{%
	\begin{tabular} {l | cc |cc}
		\toprule 
		\multicolumn{1}{c}  {}& \multicolumn{2}{c}  {1-shot}& \multicolumn{2}{c}  {5-shot}\\
		 &BC5CDR &NCBI & BC5CDR&NCBI   \\ 
          \midrule
        
         P1&   92.84 &   81.29 &    93.33  & 84.06  \\
         P2&   \textbf{99.06}  & \textbf{ 95.92 }   &   \textbf{99.06}     &   \textbf{95.92}     \\ 
        \bottomrule            
	\end{tabular}
 }
  \caption{ Few-shot named entity recognition results in BC5CDR, and NCBI. In P1, the model is the same as the PromptNER approach, while in P2, it assumes that the PromptNER is aware of all entity mentions present in each sentence.
  }
  \vspace{+2mm}
\label{con:Model_assume_all_entity}
\end{table}

The NER process consists of two crucial steps: entity recognition and entity labeling. In our study, as shown in Table~\ref{con:Model_assume_all_entity}, we have demonstrated that LLMs exhibit a strong capability for accurately labeling entities by comparing the results of P1 and P2. However, we have observed that recognizing the surface name or identifying the explicit mentions of relevant entities within a sentence can be challenging for LLMs. This particular finding is being left for further research.

\begin{figure*}[htbp]
	\centering
	\includegraphics[width=1.8\columnwidth]{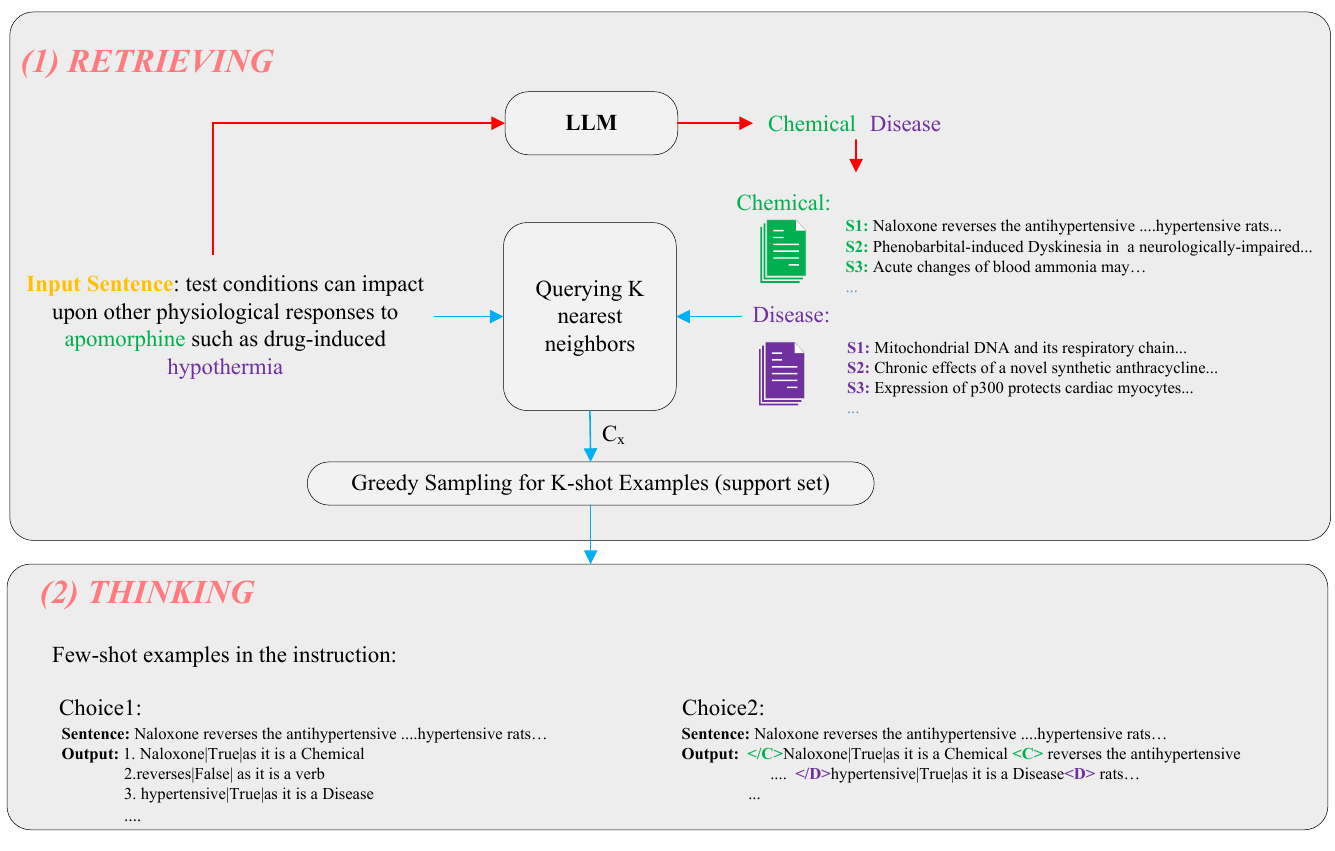} 
	\caption{The overview about our framework RT.}
	\label{con:RT}
\end{figure*}

\section{Our Method}

Based on the findings mentioned above, we introduce a novel approach called RT (Retrieving and Thinking) and present it in Figure~\ref{con:RT}. The RT method comprises two primary steps: (1) retrieving the most pertinent examples for the given test sentence, which are incorporated as part of the instruction in the ICL. This step is accomplished through the process of \textbf{Retrieving}. (2) guiding LLM to recognize the entity gradually, demonstrating this progression as \textbf{Thinking}. In the following sections, we provide a comprehensive explanation of each component.

\begin{table*}[ht]
	\centering
 \renewcommand\arraystretch{1.3}
\resizebox{0.7\textwidth}{!}{%
	\begin{tabular} {l | cc |cc}
		\toprule 
		 \multicolumn{1}{c}  {}& \multicolumn{2}{c}  {1-shot}& \multicolumn{2}{c}  {5-shot}\\
		Approach &BC5CDR &NCBI & BC5CDR&NCBI   \\ 
          \midrule
        PromptNER~\cite{ashok2023promptner}&   92.84 &   81.29 &    \textbf{93.33}  & 84.06  \\
        GPT-NER*~\cite{wang2023gpt}&  91.28  &  90.44   &  91.56  & 90.44  \\
        GPT-NER*(self-ve)~\cite{wang2023gpt}& 91.28   &  90.79  & 91.63  &   90.67 \\
        RT (Our method)&   \textbf{93.50} &   \textbf{91.56}  &  93.26    &  \textbf{91.76}    \\
        \bottomrule            
	\end{tabular}
 }
  \caption{ Few-shot medical NER results in BC5CDR, and NCBI. self-ve refers to  self-verification.
  }
  \vspace{+2mm}
\label{con:Our_Model_performance}
\end{table*}

\subsection{Retrieving}

As depicted in Figure~\ref{con:RT}, the process of obtaining relevant examples involves two distinct steps. The first step is represented by the red line in Figure~\ref{con:RT}, while the second step is indicated by the blue line in Figure~\ref{con:RT}.

In the initial step, we employ a random selection of LLMs to determine the labels that correspond to the test sentence. For this purpose, we utilize the Vanilla ICL method proposed by ~\cite{li2023w}. In Figure~\ref{con:RT}, the relevant labels about the input sentence are \textit{Chemical} and \textit{Disease}. Subsequently, we proceed to obtain candidate sentences (instances) for each label $l$ from the development dataset. This is done by considering sentences $S$ that contain an entity labeled as $l$. So the candidate sentences for label $l$ can be represented as $l: \{S_1^l, S_2^l,...,S_n^l\}$,  where $n$ represents the number of candidate sentences for a given label $l$.

In the subsequent step, we begin by embedding the input sentence $X$ into its corresponding sentence embedding $\mathbf{X}$. Additionally, we generate embeddings for the candidate sentences, denoted as ${\mathbf{S_1^l}, \mathbf{S_2^l}, ..., \mathbf{S_n^l}}$.
%%%%%%%%%
Next, we calculate the similarity between the sentence embedding $\mathbf{X}$ and each candidate embedding $\mathbf{S_n^l}$ for every label $l$. Based on these similarity scores, we select the top $K$ most similar candidate sentences and add them to the retrieved sentence set $C_x$ for the input sentence.
%%%%%%%%%%%%
To ensure that each label has a limited number of entities (either one or five), we utilize the Greedy Sampling approach to select the final examples from the retrieved sentence set.

\subsection{Thinking}

In the implementation of this step, we incorporate thinking examples within the instruction, as depicted in Figure~\ref{con:RT}. These thinking examples aid the Language Labeling Model (LLM) in gradually recognizing medical entities. Specifically, following a similar approach as PromptNER~\cite{ashok2023promptner}, the first step involves making a judgment as to whether a particular entity belongs to the candidate labels. Subsequently, in the second step, the LLM provides a rationale explaining why the entity is associated with a specific label.

\subsection{Main Results}

Table~\ref{con:Our_Model_performance} presents the experimental results of various approaches based on the Micro-F1. We have the following observations: (1) RT has an improved
performance on BC5CDR-1-shot, NCBI-1-shot, NCBI-5-shot over the best baselines GPT-NER and PromptNER,
and is almost neck to neck on BC5CDR-5-shot.
Overall, RT outperforms or achieves comparable exact-match F1 performances to the
other methods, demonstrating the effectiveness of our proposed method.
(2)  Our method evaluates the effectiveness by retrieving a good example and guiding the LLM to extract the entity by the right logic. (3) In our method, we observed that RT achieves better performance on the NCBI dataset when utilizing only positive examples (Choice 2 in Figure~\ref{con:RT}). Conversely, for the BC5CDR dataset, Choice 1 proves to be more effective. These findings indicate that different datasets have their own distinct examples that yield optimal performance. Therefore, it is crucial to adapt the choice of examples based on the specific dataset being addressed. By understanding and leveraging dataset-specific examples, we can enhance the performance and applicability of the RT method.

\section{Error Analysis}
\label{Error Analysis}
% We further manually analyze the reason about the RT cannot get 100\%  medical entity recognition  on the test set of BC5CDR and NCBI, using the 1-shot, 5-shot support set separately,  and summarize them into the following categories (examples for each category are shown in Tabel \ref{con:RT_error_prediction}.

We conducted an additional manual analysis to determine the reasons why RT fails to achieve 100\% accuracy in medical entity recognition on the test sets of BC5CDR and NCBI. This analysis was performed separately for the 1-shot and 5-shot support sets. We have categorized the findings into different categories, and examples for each category are presented in Table \ref{con:RT_error_prediction}.

\begin{itemize}
  \item \textbf{Unable to extract entities:} Our analysis indicates that RT struggles to
  comprehend the semantic information of input sentences and accurately identify the corresponding entities. 
 
 \item \textbf{Misidentification:}  RT often struggles to recognize and generate the entire entity, instead only producing a partial representation of the gold entity.

 \item  \textbf{ Class collision:}  During our analysis, we observed that the test data is only partially annotated, which prevents the RT gets 100\%  entity recognition.
 
 \item  \textbf{Multi-label entities:}  During our analysis, we observed that  some entities have more than one label.
 
 \item  \textbf{ Unable to generate symbols:} Despite RT's ability to extract complex entities like \textit{ severe combined immunodeficiency syndrome}, it faces difficulties in generating special symbols within certain entities.  such as the gold entity is \textit{B-cell non-Hodgkin's lymphoma}, the predicted entity is \textit{B-cell non-Hodgkin\textbackslash"s lymphoma}.

\item \textbf{Wrong Template Prediction:} Despite being provided with output templates and examples, RT still generates a considerable number of output errors that are not based on the provided template. 

\end{itemize}

\begin{table}[ht]
	\centering
		
	\renewcommand\arraystretch{1.3}
 
	\scalebox{0.8}{
	\begin{tabular} {lll}
		\hline 
		\textbf{Dataset}&\multicolumn{2}{l}{\textbf{Unable to extract entities:}}  \\ 
		\hline

		\multirow{3}*{NCBI}&Test sentence:&\makecell[l]{Myotonic dystrophy protein \\ kinase is involved in \\ the modulation...}\\
		%~&Cosine-Map&0.097&0.212&0.134&  0.042& 0.084&0.198&0.109&0.036\\
		~&\textcolor{teal}{gold entities} & \textcolor{teal}{\makecell[l]{Myotonic dystrophy,\\ muscular disorder,\\DM}} \\
        ~&\textcolor{orange}{predicted entities} & \textcolor{orange}{\makecell[l]{DM}} \\

      \hline
  \textbf{}&\multicolumn{2}{l}{\textbf{Misidentification:}}    \\ 
\hline
	\multirow{2}*{NCBI}&\textcolor{teal}{gold entities} & \textcolor{teal}{\makecell[l]{breast malignancies} }\\
		%~&Cosine-Map&0.097&0.212&0.134&  0.042& 0.084&0.198&0.109&0.036\\
		~&\textcolor{orange}{predicted entity:} & \textcolor{orange}{\makecell[l]{breast} }\\
   \hline

\multirow{2}*{NCBI}&\textcolor{teal}{gold entities} & \textcolor{teal}{\makecell[l]{non-familial cancers"} }\\
		%~&Cosine-Map&0.097&0.212&0.134&  0.042& 0.084&0.198&0.109&0.036\\
		~&\textcolor{orange}{predicted entity:} & \textcolor{orange}{\makecell[l]{cancers} }\\
   \hline

\multirow{2}*{BC5CDR}&\textcolor{teal}{gold entities} & \textcolor{teal}{\makecell[l]{ate-onset scleroderma\\ renal crisis} }\\
		%~&Cosine-Map&0.097&0.212&0.134&  0.042& 0.084&0.198&0.109&0.036\\
		~&\textcolor{orange}{predicted entity:} & \textcolor{orange}{\makecell[l]{scleroderma renal crisis } }\\
   \hline

      \hline
    \textbf{}&\multicolumn{2}{l}{\textbf{Class collision:}}    \\ 
    \hline
	\multirow{1}*{BC5CDR}&\textcolor{blue}{error description:} & \textcolor{blue}{\makecell[l]{acute cardiotoxicity\\ is not annotated } }\\
   \hline
  \textbf{}&\multicolumn{2}{l}{\textbf{Multi-label entities:}}    \\ 
    \hline
	\multirow{1}*{NCBI}&\textcolor{blue}{situation description:} & \textcolor{blue}{\makecell[l]{the entity \textit{DM} has \\two labels \\SpecificDisease and Modifier } }\\
   \hline
\textbf{}&\multicolumn{2}{l}{\textbf{Unable to generate symbols:}}    \\ 
    \hline
	\multirow{2}*{BC5CDR}&\textcolor{teal}{gold entity:} & \textcolor{teal}{\makecell[l]{B-cell non-Hodgkin's \\ lymphoma} }\\
 ~&\textcolor{orange}{predicted entity:} & \textcolor{orange}{\makecell[l]{B-cell non-Hodgkin\textbackslash"s \\ lymphoma } }\\
   \hline
\textbf{}&\multicolumn{2}{l}{\textbf{Wrong template prediction:}}    \\ 
    \hline
	\multirow{2}*{NCBI}&\textcolor{teal}{gold template:} & \textcolor{teal}{\makecell[l]{</M>WD|True| \\as it is Modifier<M>} }\\
 ~&\textcolor{orange}{predicted template:} & \textcolor{orange}{\makecell[l]{</M>WD|True|as it is \\ Modifier<M> cannot be \\ homologous to </M>"} }\\
   \hline

\hline
	\end{tabular}}
	\caption{The examples of the reason about  RT cannot get 100\% medical entity recognition.}
 
\label{con:RT_error_prediction}
\end{table}

\section{Conclusion}
We have conducted an extensive empirical study comparing 16 SLMs and LLMs on two open-domain medical NER datasets. We show that LLMs are good few-shot medical NER extractors with good examples and  reasonable extraction logical. Building on some important findings, we propose a RT (Retrieving and Thinking) method to improve the performance of medical NER. 
We also found it is true that LLMs may encounter difficulties in extracting entities with 100\% accuracy. There are several reasons for this, including misidentification, wrong template prediction, etc.

\section*{Acknowledgements}
This work was supported by the National Institutes of Health’s National Center for Complementary and Integrative Health grant number R01AT009457 and National Institute on Aging grant number R01AG078154. The content is solely the responsibility of the authors and does not represent the official views of the National Institutes of Health.

\bibliography{anthology}
\bibliographystyle{acl_natbib}

\end{document}